**Sampling Limits for Electron Tomography with Sparsity-exploiting Reconstructions**

Yi Jiang[1], Elliot Padgett[2], Robert Hovden[3], David A. Muller[2,4]

[1] *Department of Physics, Cornell University, Ithaca, NY 14853*

[2] *School of Applied & Engineering Physics, Cornell University, Ithaca, NY 14853*

[3] *Department of Materials Science and Engineering, University of Michigan, Ann Arbor, MI 48109*

[4] *Kavli Institute at Cornell for Nanoscale Science, Cornell University, Ithaca, NY 14853*

Electron tomography (ET) has become a standard technique for 3D characterization of materials at the nano-scale. Traditional reconstruction algorithms such as weighted back projection suffer from disruptive artifacts with insufficient projections. Popularized by compressed sensing, sparsity-exploiting algorithms have been applied to experimental ET data and show promise for improving reconstruction quality or reducing the total beam dose applied to a specimen. Nevertheless, theoretical bounds for these methods have been less explored in the context of ET applications. Here, we perform numerical simulations to investigate performance of $\ell_1$-norm and total-variation (TV) minimization under various imaging conditions. From 36,100 different simulated structures, our results show specimens with more complex structures generally require more projections for exact reconstruction. However, once sufficient data is acquired, dividing the beam dose over more projections provides no improvements—analogous to the traditional dose-fraction theorem. Moreover, a limited tilt range of ±75° or less can result in distorting artifacts in sparsity-exploiting reconstructions. The influence of optimization parameters on reconstructions is also discussed.

## 1. Introduction

Electron tomography (ET) attempts to reconstruct the 3D structure of physical and biological materials from an angular range of 2D images collected by a (scanning) transmission electron microscope ((S)TEM) (De Rosier and Klug, 1968; Hart, 1968; Gordon et al., 1970; Radermacher, 1988; Baumeister et al., 1999; Koster et al., 2000; Koguchi et al., 2001; Midgley et al., 2001). The set of images is often referred to as a tilt series and modeled as projections of the original object. Unfortunately, in typical ET experiments, radiation damage, contamination, and acquisition time limit the signal-to-noise ratio (SNR) and the number of projections (ca. ~70-140). Furthermore, specimen and stage geometry usually restrict the tilt range (ca. ±70°), leaving a large missing wedge of information in Fourier space. Consequently, conventional reconstruction algorithms, such as weighted back projection, that only make use of measured data suffer from elongation and blurring artifacts that are disruptive to accurate characterizations.

Recently, there is a growing interest in developing reconstruction techniques that incorporate additional prior knowledge about the specimen (Batenburg et al., 2009; Miao et al., 2005; Saghi et al., 2011). Inspired by the field of compressed sensing, a majority of these methods (Saghi et al., 2011; B. Goris et al., 2012; Bart Goris et al., 2012; Leary et al., 2013; Goris et al., 2013) exploit the notion of image sparsity and obtain reconstructions via minimizing the $\ell_1$-norm of the object vector in some domain. Unlike back projection, sparsity-exploiting methods have considerable flexibility in designing reconstructions based on users' assumptions about the specimen as well as the desired utility of the reconstruction (Han et al., 2012). A typical form of the optimization problem can be written as

$$\underbrace{\min\|D(x)\|_1}_{\text{Objective Function}} \text{ s.t. } \underbrace{\|Ax - b\|_2 \leq \varepsilon,}_{\text{Data Constraint}}$$

where the Objective Function term has Reconstruction ($x$), and the Data Constraint term has Measurement Matrix ($A$), Data ($b$), and Data-Tolerance Parameter ($\varepsilon$).

where $x$ and $b$ represent the reconstructed image and measured data. $A$ is referred to as the "*measurement matrix*", which *models* the experimental imaging process and depends on sampling schemes such as pixel/voxel size, the number of projections and tilt range. A scalar "*data-tolerance parameter*" $\varepsilon$ is introduced to accommodate inconsistencies between data ($b$) and imaging model ($Ax$). Higher SNR data generally allows a smaller $\varepsilon$ to be used because there are fewer discrepancies between the measured data and the reconstruction model. The function $D(x)$ transforms the reconstruction to a domain in which the object is assumed to be sparse. Two most popular transformations are identity ($D(x) = x$) and gradient magnitude ($D(x) = \|\nabla x\|_2$). The $\ell_1$-norm of the gradient-magnitude image is also known as the TV-norm ($\|x\|_{\text{TV}}$) (Rudin et al., 1992) and has been used in image processing and reconstruction algorithms for over two decades (Vogel and Oman, 1996; Panin et al., 1999). Furthermore, for STEM tomography, it is also beneficial to enforce additional constraint that restricts pixels/voxels to be non-negative.

To date, several experimental works have demonstrated that sparsity-exploring methods can reduce artifacts and improve overall reconstruction quality (Saghi et al., 2011; B. Goris et al., 2012; Leary et al., 2013) for under-sampled data. Nevertheless, theoretical limitations of such algorithms are seldom discussed in the context of ET, especially when and how they fail. In essence, the reconstruction (a solution of the optimization problem) can be interpreted as a multivariable function that depends on $A$, $b$, and $\varepsilon$. Different experimental conditions, such as sampling scheme or data quality, and the data-tolerance parameter can lead to significantly different reconstructions. It is generally onerous, if not impossible, to explore the entire parameter space when characterizing optimization-based reconstructions. Thus, in this work, we carry out extensive simulation studies of both $\|x\|_1$ and $\|x\|_{\text{TV}}$ minimization techniques and investigate four key parameters that are of practical interests: number of projections, data-tolerance parameter, data noise, and tilt range. Our results demonstrate some fundamental behaviors of sparsity-exploiting methods:

1. The number of projections required for exact reconstruction increases with specimen complexity.

2. In the presence of Poisson noise, the quality of sparsity-exploiting reconstructions degrades quickly. With a sufficient number of projections, the root-mean-square error (RMSE) of the reconstruction depends only on the total electron counts (i.e. dose). Using more projections with lower SNR has insignificant influence on the reconstruction. This resembles the traditional dose-fraction theorem for weighted back projection (Saxberg and Saxton, 1981; Hegerl and Hoppe, 1976).

3. Sparsity-exploiting reconstructions suffer from distorting artifacts when the tilt range is less than ±75°.

4. The data-tolerance parameter also has significant impact on reconstructions. For TV minimization, small $\varepsilon$ can produce noisy artifacts, while large $\varepsilon$ results in over-smoothed reconstructions.

These results provide basic insights on data acquisition and reconstructions for ET. Because the number of projections required for sparsity-exploiting reconstruction is specimen-dependent, one should be cautious when reducing the number of projections or tilt range in practical experiments. With proper choice of the optimization parameter, however, sparsity-exploiting reconstructions are robust to a small missing wedge (~30°, i.e. tilt range of greater than ±75°) and follow the traditional dose-fraction theorem. Increasing data SNR generally improves reconstruction quality.

**2. Background**

2.1 Image model for ET

Most optimization-based reconstruction methods are built upon the *discrete-to-discrete model* (Barrett and Myers, 2003) in which both image ($x$) and data ($b$) are represented as vectors and related via the measurement matrix as $Ax = b$ (Eq.1). In this model, tomographic reconstruction is an inverse problem of solving x for a given system of linear equations.

In ET, each measurement is often interpreted as line integrals (i.e. a projection) across the specimen(Hoppe and Hegerl, 1980). If data is *ideal*, that is, $b = Ax$, one can define a *sufficient projection number* as the smallest number of projections that gives a full rank measurement matrix (Jorgensen et al., 2013), which guarantees that Eq.1 has a unique solution. Due to experimental limitations, the sufficient projection number is rarely achieved in practice. For instance, in order to reconstruct a 512 × 512 image from data with 1024 measurements in each projection, one needs to record at least 256 projections (the exact number depends on how $A$ is constructed). To overcome data insufficiency, additional knowledge beyond Eq. 1 needs to be incorporated into the imaging model so that the reconstruction is closer to the actual specimen.

2.2 Sparse model and compressed sensing

The notion of sparsity has become widely used in optimization-based reconstruction techniques. Mathematically, sparsity refers to the number non-zero elements in a vector (also known as the $\ell_0$-norm). A sparse model assumes only a small fraction of the elements in the image vector ($x$), or some transform of it, are non-zero. Because direct $\ell_0$-norm minimization is NP-hard (Garey and Johnson, 1979), alternative approaches have been developed to approximate sparse solutions. In 2002, Li *et al* (Li et al., 2002) used an $\ell_1$-norm minimization method to reconstruct sparse blood vessels from 15 X-ray computed tomography (CT) projections. The idea of $\ell_1$-norm minimization is further reinforced by the work of Candes *et al* (Candes et al., 2006), which proved that if the measurement matrix satisfies a restricted isometry property (RIP), it is highly probable the sparsest solution and minimal $\ell_1$-norm solution are equivalent, given there are enough non-zero measurements. This result led to the field of compressed sensing (Candes et al., 2006; Donoho, 2006) and re-invigorated developments in optimization-based methods.

Despite the popularity of compressed sensing, its theoretical conclusions are of limited use in practice. It is well known that compressed sensing typically favors dense and random measurement matrices (Jakob S. Jørgensen et al., 2014). In ET or X-ray CT, on the other hand, measurement matrices are much more sparse and structured (See Figure 1 in ref. (Jakob Sauer Jørgensen et al., 2014)). Studies using radon transforms have shown not only the existence of sparse vectors that cannot be reconstructed by $\ell_1$-norm minimization (Pustelnik et al., 2012), but RIP-based guarantee only holds for extremely sparse vectors (Sidky et al., 2010). Moreover, it is computationally intractable (NP-hard) to examine whether a measurement matrix satisfies the properties required by compressed sensing (Tillmann and Pfetsch, 2014). Without theoretical guarantees, it is imperative to carry out simulation studies using an ensemble of objects with well-defined features to understand the recoverability of any algorithm.

**3. "Phase diagram" analysis for $\ell_1$-norm minimization**

3.1 Simulation Design

In this section, we perform a "phase diagram" analysis to study the recoverability of $\ell_1$-norm minimization method at various imaging conditions. Adapting the work by Jørgensen *et al* (Jørgensen, 2015; Jørgensen and Sidky, 2015), we establish an average-case relation between image sparsity and the number of projections needed for exact reconstruction. In Figure 1, the simulation results are summarized as a function of the percentage of non-zero pixels ($k$) of the object intensity and relative sampling ($\mu$), which is defined as the ratio of the number of projections to the sufficient projection number. For each pair of ($k, \mu$) in this phase space, we generate 100 semi-realistic test objects with similar complexity. Details of object generation are summarized in supporting information and the source code is included in the tomography software *tomviz* (www.tomviz.org) (Jiang et al., 2016b). For ideal data, a reconstruction is obtained by solving the basis pursuit optimization problem: $\min\|\mathbf{x}\|_1 \; s.t. \; Ax = b$ (van den Berg and Friedlander, 2008). A total of 36,100 different structures are used in the phase diagram and we report the percentage of "accurate" reconstructions whose RMSE are less than 0.05 in Fig. 1(a).

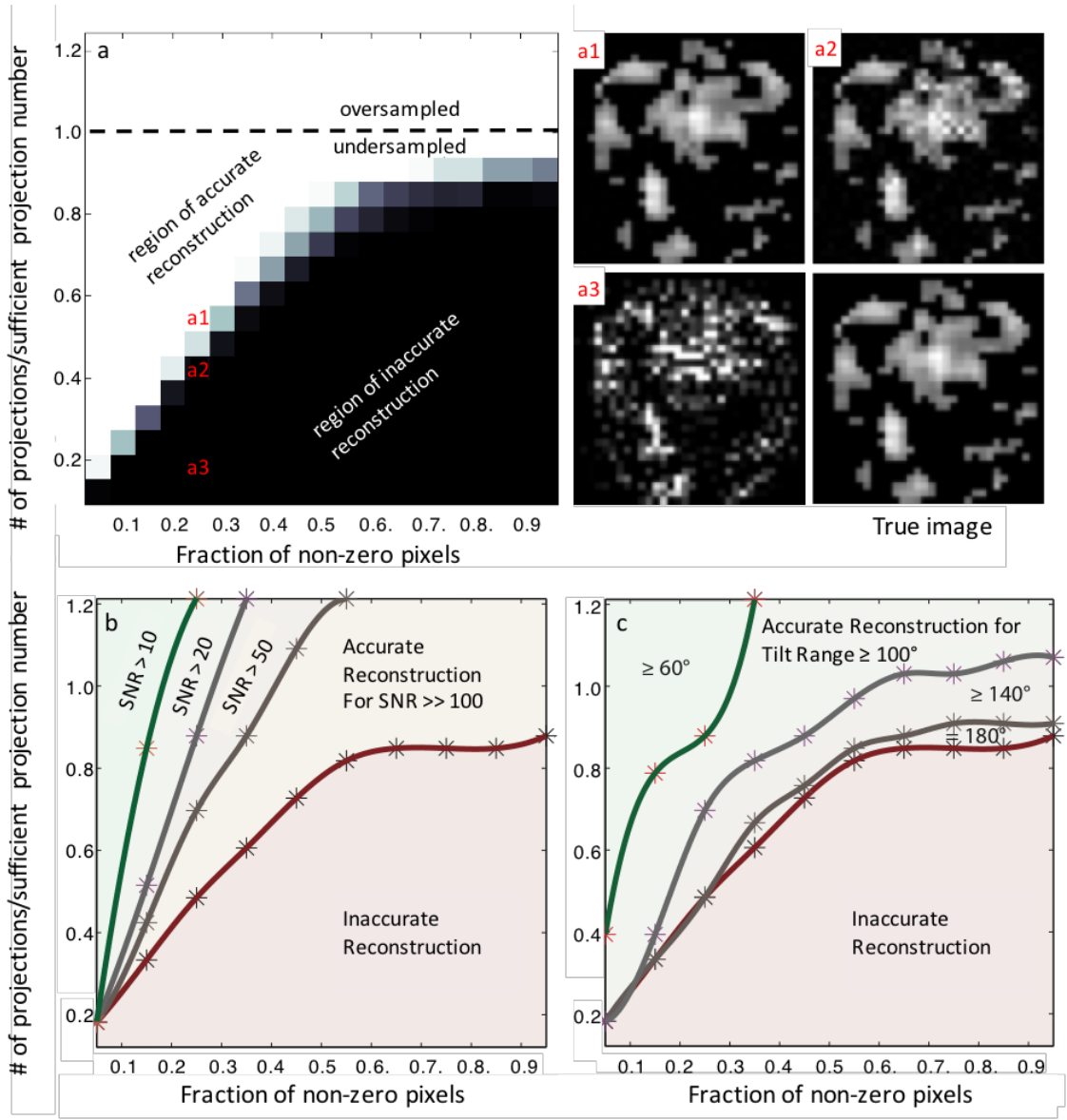

**Figure 1**: Phase diagram of $\ell_1$-norm minimization reconstruction. a.) For ideal data, the percentage of accurate reconstructions as a function of percentage of non-zero pixels and relative sampling (number of projection/sufficient projection number). There is a sharp transition from the region where all the test images are accurately reconstructed (RMSE≤0.05) to the region of zero recovery rates. a1-a3.) Three reconstructions with different numbers of projections. In the region of accurate reconstruction, the reconstruction matches the true image. As relative sampling decreases, noticeable artifacts start to appear. b.) Transition boundaries of noisy data for different signal-to-noise ratios. The presence of Poisson noise greatly reduces the region of accurate

reconstruction. c.) Transition boundaries vs. tilt range. The boundary for a ±70° tilt range is close to that of the full tilt range, but the method has very limited success when the tilt range is below ±50°.

In general, $\ell_1$-norm minimization reconstructions have "Crowther-like" (Crowther et al., 1970) behavior: as the number of non-zero pixels increases, the algorithm requires more projections to maintain the same level of numerical accuracy. Even sparse specimens of modest complexity contain many artifacts without sufficient projections (Fig.1 a2, a3). Nevertheless, exact reconstruction is possible with more projections (Fig.1 a1). Our simulation demonstrates that for the ET measurement matrix, $\ell_1$-norm minimization indeed leads to the true image even when the number of projections is smaller than sufficient projection number. More studies about the phase diagram approach, such as the dependence on image class, uniqueness tests, and TV minimization reconstructions can be found in ref. (Jakob Sauer Jørgensen et al., 2014; Jørgensen, 2015; Jørgensen and Sidky, 2015).

The transition boundary in Figure 1(a) sets a benchmark for studying more realistic systems. In the presence of Poisson noise, we modify the data constraint to $\|Ax - b\|_2 \leq \varepsilon$ and chose the data-tolerance parameter that gives the smallest RMSE. As shown in Figure 1b, because the measurement matrix is ill-conditioned, the region of accurate reconstruction is significantly reduced, even for relatively high SNR (50:1) data. Moreover, Figure 1c reveals that the "missing wedge" problem also shifts up the transition boundary. The deviation is relatively small at for maximum tilts ranges ±70° and higher, but the method has very limited success when the tilt range is below ±50°.

**4. Dose-allocation study of TV minimization**

4.1 Simulation Design

In this section, we explore various limitations of the TV minimization method. For simplicity, we consider a single tilt axis geometry so that planes that are perpendicular to the tilt axis can be reconstructed separately. The test image (with 256 × 256 pixels) is constructed to resemble an experimental annular dark-field STEM tomographic reconstruction of a fuel cell catalyst with platinum nanoparticles on a porous carbon support. This system is of current research interest (Liu et al., 2008; Yu et al., 2012) and presents challenges for ET characterization due to the large platinum/carbon contrast ratio and small features of interest. The image has a sparse gradient-magnitude image with only 3.3% of all pixels being non-zero. The intensity of carbon and platinum are 0.03 and 0.75. To focus on the structure of carbon support, all images are displayed in a tight contrast window of [0.01, 0.05], along with a zoomed-in image to highlight the porous structure (Fig. 2).

Many optimization algorithms are available for solving the constraint-TV minimization problems. Here, we implement a heuristic approach that adaptively uses steepest descent and projection-on-to-convex-set (Gordon et al., 1970; Youla and Webb, 1982; Combettes, 1993) to reduce the image TV and data residual. The algorithm is often referred as ASD-POCS (Sidky and Pan, 2008) and has been used in various large-scale tomography reconstruction problems (Sidky and Pan, 2008; Bian et al., 2010; Han et al., 2012; Bian et al., 2014; Han et al., 2015). A detailed description and verification of the algorithm are provided in supplemental material.

To ensure the algorithm reaches an optimal solution, a scalar quantity— referred to "cosine alpha" ($c_\alpha$), which equals -1 only at optimal solutions (Sidky and Pan, 2008), is calculated at each iteration. It has been shown that $c_\alpha$ is an effective metric in monitoring the convergence of the algorithm (Sidky and Pan, 2008; Bian et al., 2010; Han et al., 2012; Bian et al., 2014; Han et al., 2015). Moreover, as discussed in ref.(Sidky and Pan, 2008), unless ε is so large that the feasible solution set contains a flat image (with zero TV), the optimal solution's data residual ($E(x)$) is always equal to the data-tolerance parameter ε. Due to finite number of iterations and computer precision, we terminate reconstructions when $c_\alpha \leq -0.95$ and $\left|\frac{E(x)-\varepsilon}{\varepsilon}\right| \leq 10^{-4}$. This practical convergence check guarantees that our reconstructions numerically and visually converge to the optimal solutions (see supporting Fig. S2).

4.2 Influence of number of projections

We first use ideal data to investigate how the number of projections influences a TV minimization reconstruction. Fig.2 shows reconstructions of the Pt/C reference image using 5, 10, 20, 30, and 40 projections that evenly cover the full tilt range (i.e. no missing wedge) and has 256 measurements in each projection. Although the simulated data is error-free, the data residual persist due to computer precision and a finite number of iterations. We therefore set the data-tolerance parameter ($\varepsilon$) to $10^{-5}$ for all reconstructions. A smaller $\varepsilon$ has an insignificant influence on the results (see supporting Fig. S3).

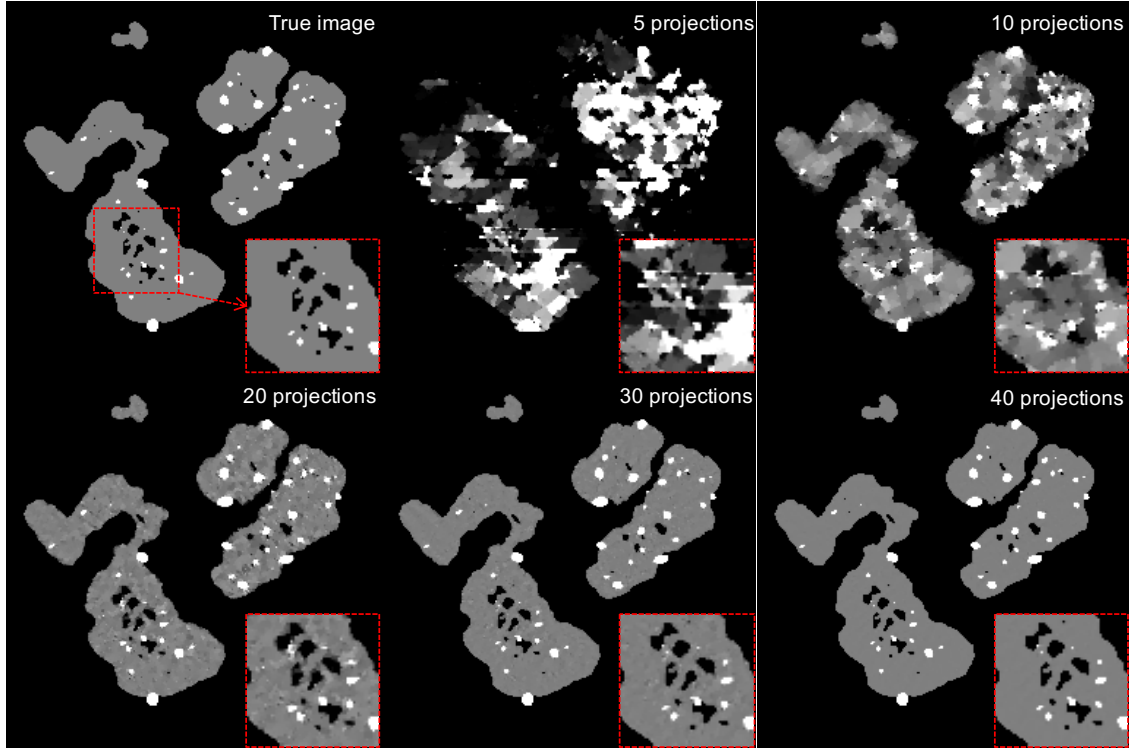

**Figure 2:** TV minimization reconstructions of the Pt/C test image using ideal data with different number of projections ($N_p$). Without sufficient data ($N_p<10$), TV reconstructions suffer from dramatic artifacts. The results improve as $N_p$ increases and become almost visually indistinguishable from the true image at $N_p=40$. The box at the bottom right corner is a zoom-in of region indicated in the true image.

With only 5 projections, the reconstruction is over-simplified and contains streaky artifacts. These artifacts are the result of under-sampling, where few constraints are provided in the measurement matrix and thus the feasible solution set includes reconstructions with much lower TV than the true image. As the number of projections increases, data constraints become stronger and more features are imposed to the image. With 20 projections, the porous structures are visually close to the true image, but the carbon support contains some intensity variations, which diminishes when more projections are used.

It is worth noting that with 40 projections, the number of non-zero elements in the data vector (8164) is about 3.8 times the sparsity of the reference image's gradient-magnitude image (2153). This agrees with the phase diagram analysis for TV minimization reported in (Jørgensen and Sidky, 2015) and could leads to a heuristic rule of thumb for minimum sampling requirement on similar specimen. Experimental data that contains noise and other sources of errors requires more projections than the theoretically lower bound, as demonstrated in sections 4.3.

4.3 Influence of data-tolerance parameter (ε)

When the measured data (*b*) are inconsistent with the image model (*Ax*), the choice of a data-tolerance parameter can have a significant impact on reconstructed images. To demonstrate this, we generate 60 projections that are corrupted with Poisson noise. The total electron counts collected by the detector ($N_e$) is set to $4\times10^4$ (the SNR is about 25.8). The tilt range is ±90° and each projection has 256 measurements. Fig. 3 shows RMSE of reconstructions as a function of ε, along with selected tomograms for visual inspection.

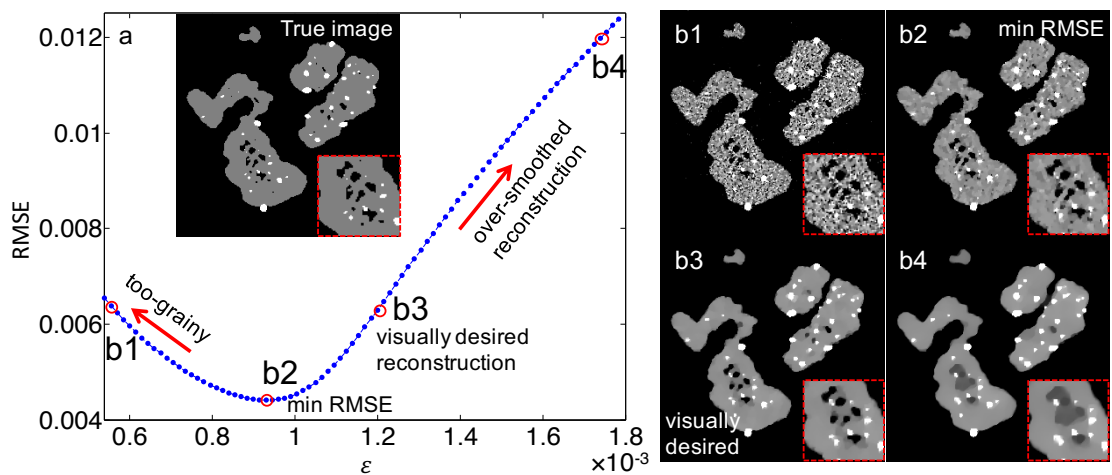

**Figure 3:** Influence of the data-tolerance parameter (*ε*) on the TV reconstructions. The Pt/C test image is reconstructed from a noisy dataset with a wide range of *ε*. a.) Reconstruction RMSE vs. data-tolerance parameter. If *ε* is too small, reconstructions are filled with grainy artifacts (b1). Larger *ε* results in over-smoothed images (b4). Due to the high intensity of Pt particles, the optimal-RMSE reconstruction (b2) is different from the reconstruction that's visually more similar to the true image (b3).

With small *ε*, the data constraint is relatively strong and limits the feasible solutions. As a result, TV minimization has insignificant influence and reconstructions are filled with grainy artifacts (Fig.3 b1). When the constraint is relaxed (by increasing *ε*), TV minimization starts to have a greater effect in removing small speckles and thus reducing RMSE. On the other hand, if *ε* is too large, reconstructions become over-smoothed and lose resolution (Fig.3 b4), causing RMSE to rise again.

Because the curve of reconstruction RMSE vs. *ε* is convex, we define the "optimal-RMSE" as the smallest error achievable for a given dataset. Note that since the reference image has large area of low intensity pixels yet RMSE is dominated by high intensity pixels, the optimal-RMSE reconstruction still contains some speckle artifacts in the carbon support (Fig.3 b2). A visually more appealing reconstruction (Fig.3 b3) could be obtained with larger *ε*. In practice, different optimization parameters can be selected based on different applications. Here, to provide a general sense of how TV reconstructions change under different conditions and avoid any bias in visual inspection, we report optimal-RMSE reconstructions in section 4.3 and 4.4.

4.4 Influence of Poisson noise

In this section, we investigate the effect of noise level and the dose-allocation problem: for a fixed number of electron counts, is it more beneficial to use more projections with less SNR for reconstruction or vice versa? The traditional dose-fraction theorem(Saxberg and Saxton, 1981), which is derived using weighted back projection, shows that with sufficient projections, the reconstruction RMSE only depends on the total electron counts. For optimization-based methods, the trade-off lies between the measurement matrix $A$, which reflects data insufficiency, and the data vector $b$, which reflects data inconsistency.

Fig.4 shows reconstructions from various total measured electrons (i.e. dose). For each dose, we generate tilt series with different numbers of projections. The tilt range is ±90° and there are 256 measurements per projection. Each dataset is reconstructed with a wide range of the data-tolerance parameter $\varepsilon$, and we choose the parameter that results in the smallest RMS reconstruction error. The optimal-RMSE as a function of projections is shown in Fig.4 a.

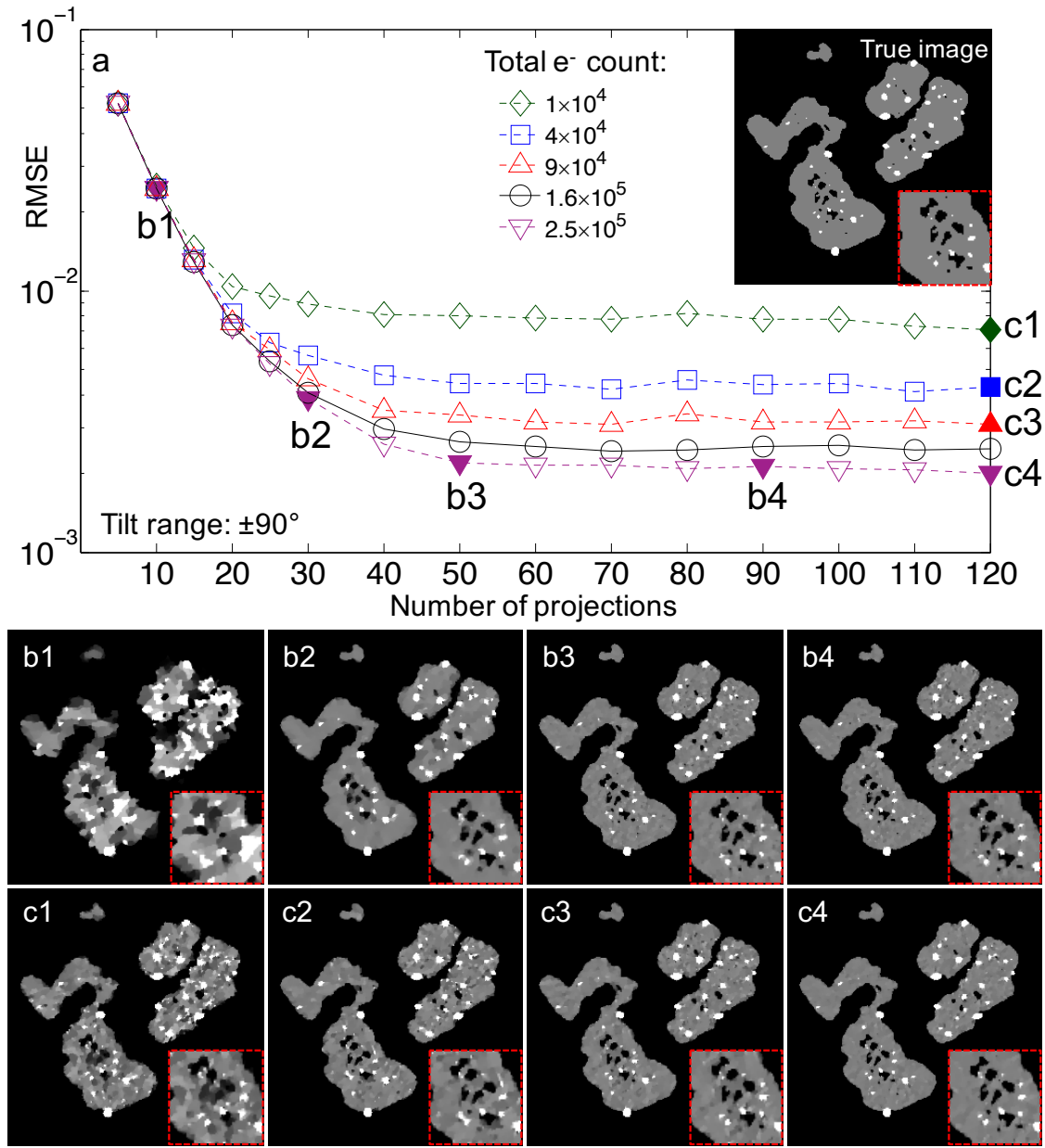

**Figure 4:** Influence of Poisson noise on TV minimization reconstructions of the Pt/C test object. The tilt range is fixed at ±90°. a.) Optimal reconstruction RMSE vs. number of projections. Each curve represents different fixed total electron counts ($N_e$). b1-b4.) Reconstructions from different number of projections at same $N_e$. c1-c4.) Reconstructions from the same number of projections with different SNR.

When the number of projections is fewer than 50, the SNR of the data is relatively high and thus the reconstruction artifacts in this region are mainly due to insufficient data, as shown in Fig.4 b1 and b2. Between 50 and 120 projections, on the other hand, the total electron counts become more critical in determining reconstruction qualities. As demonstrated in Fig.4 c1–c4, higher doses lead to more accurate reconstructions. Similar to the traditional dose-fraction theorem, there is no significant numerical or visual improvement with more projections (Fig.4 b3, b4, c4), indicating that benefits from having stronger data constrains is canceled by the increasing errors due to data inconsistency. Our simulations with other test images and $\ell_1$-norm minimization show similar results.

4.5 Influence of limited tilt range (missing wedge)

Lastly, we demonstrate that TV minimization reconstructions are robust to limited tilt range, provided that the missing wedge is small. Using the $N_e=1.6\times10^5$ curve in Figure 4 as a reference, we generate data at the same noise level whose projection angles are evenly distributed between -θ to θ, where θ=30°, 45°, 60°, 75°. The reconstruction results are summarized in Fig. 5.

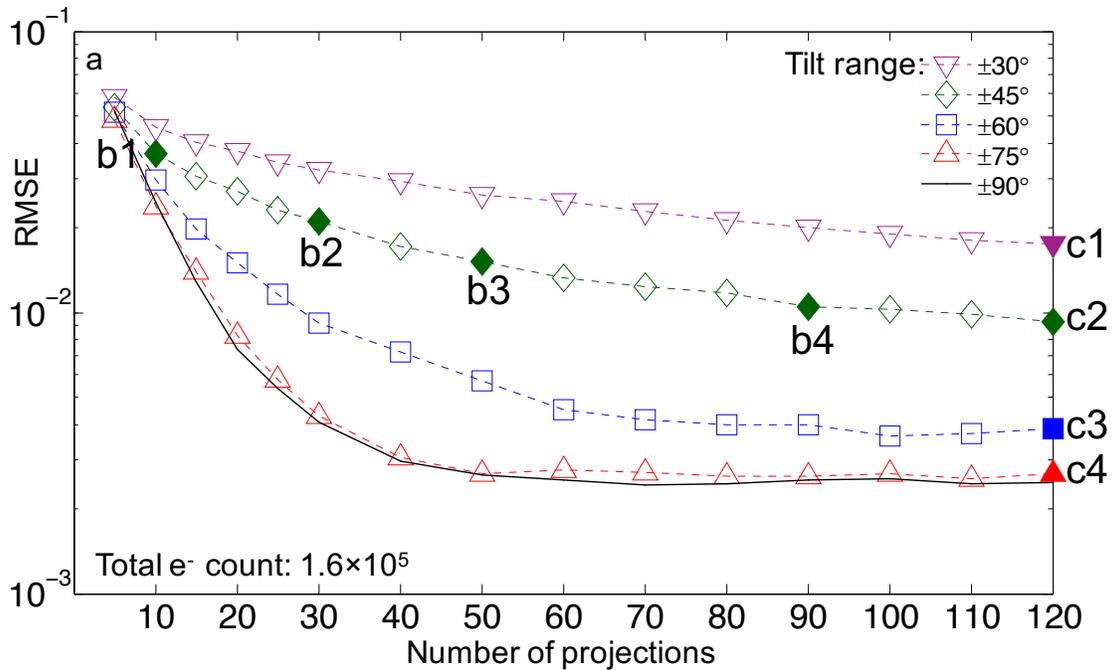
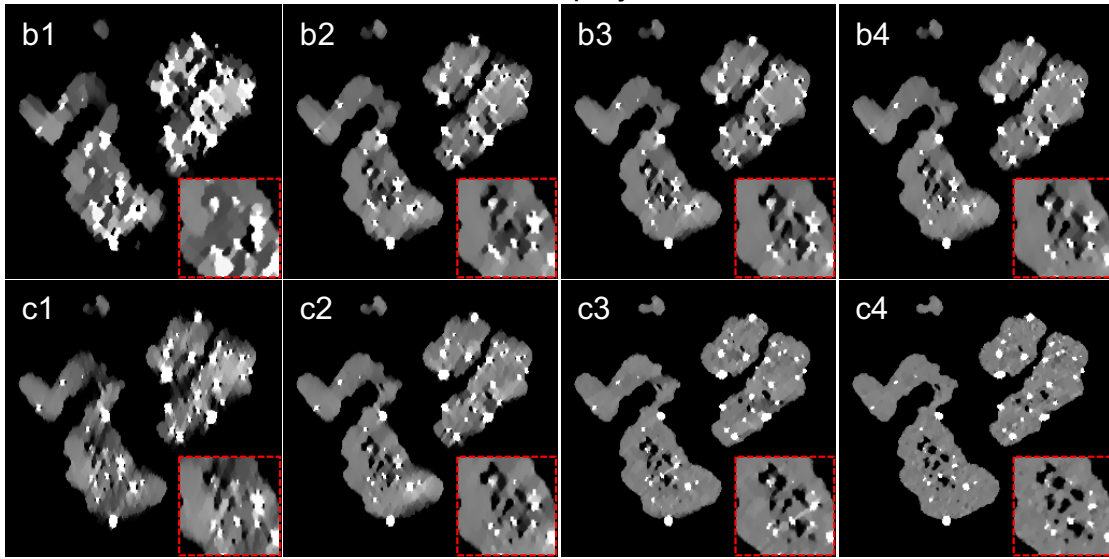

**Figure 5:** Influence of limited tilt range on TV minimization reconstructions of the Pt/C test image. The total electron counts are fixed at 1.6×10⁵. a.) Optimal reconstruction RMSE vs. number of projections. Each curve represents a different tilt range. b1-b4.) Reconstructions from different number of projections with same tilt range, as marked in (a). c1-c4.) Reconstructions from the same number of projections with different tilt range (missing wedge is vertically aligned).

When the tilt range is limited, as shown in Fig.5 c1-c3, the platinum particles in TV minimization reconstructions suffer from distortions and there are intensity variations in the carbon support. For fixed $N_e$, improvements with increasing number of projections are slower compared with full tilt range (Fig.5 b1-b4). Nevertheless, the artifacts diminish as tilt range increases and a small missing wedge (less than 30°) has a negligible influence on the reconstruction (Fig.5 c4).

**5. Discussion**

Our simulations show that sparsity-exploiting methods can achieve accurate reconstructions using a number of projections that is fewer than the sufficient projection number. For the Pt/C test image, there is marginal benefit for TV minimization to use more than 50 projections. Although this number is obtainable in most ET experiments, we want to remind readers that the amount of data needed is always specimen-dependent, as demonstrated in Fig.1 and ref. (Jakob Sauer Jørgensen et al., 2014; Jørgensen, 2015; Jørgensen and Sidky, 2015). Physical objects are often more complex and hence require more projections than most simplified test images. Another practical limitation is that specimens often have larger volumes than the field of view of reconstruction, which introduces additional artifacts to the tomogram. It's been shown that incorporating a derivative operator to the data constraint can alleviate the data truncation artifacts (Sidky et al., 2014a, 2014b).

Moreover, evaluations of reconstructions should be task-specific. For instance, based on Fig. 2, one would only need 20 projections to distinguish platinum particles from the carbon support. Fig. 3 also shows that numerically less accurate solutions looks similar to the true image. In this paper, we use RMSE and visual inspection for demonstrating how the reconstruction method responds to different imaging conditions. In practice, such assessments may be inadequate if one is interested in other utilities such as segmentation and classification. Further investigations are needed to fully characterize sparsity-exploiting algorithms.

As shown in Section 4.2, besides sampling (*A*) and data quality (*b*), another key to a successful reconstruction is the data-tolerance parameter. The results presented in section 4.3 and 4.4 are likely to change for a different parameter-selection protocol. In simulations, $\varepsilon$ is selected based on the RMSE vs. $\varepsilon$ curves since we have the full knowledge of the reference image. For experimental data, such curve cannot be made and visual inspection is often used. We have performed additional analysis by visually choosing reconstructions with more accurate carbon structure and observed that the dose-allocation principle and the robustness to small missing wedge still apply to TV minimization method. Future work should explore other parameter-selection techniques that are tailored to specific applications in ET (Jiang et al., 2016a).

Finally, we want to emphasize that we do not intend to promote the constrained $\ell_1$-norm or TV minimization above any other reconstruction technique. Instead, the purpose of the work is to investigate technical characteristics of these techniques and set the stage for forthcoming studies using experimental data and other optimization-based methods.

**Conclusion**

In this work, we have investigated the performance of sparsity-exploiting minimization methods under various conditions in ET systems. Our simulations show that the both $\ell_1$-norm and TV minimization reconstructions primarily depend on the number of projections. The amount of data required for accurate reconstruction depends on the complexity of the specimens. Moreover, the parameters in the optimization program can have significant impact to reconstructions. By choosing the data fidelity parameter ε that minimizes reconstruction RMSE, we found the numerical accuracy of TV minimization reconstructions is independent of the number of projections for fixed total electron counts, except for highly under-constrained reconstructions with very few tilts. Lastly, we demonstrated the sparsity-exploiting methods are robust to a small missing wedge of 30° or less. The results provide a general guideline on how to apply these reconstruction techniques in ET and serve as the benchmark for comparative studies with other optimization-based methods.

**Acknowledgements**

We gratefully thank Dr. Xiaochuan Pan from the University of Chicago for many helpful discussions. The tomviz project is supported by DOE Office of Science contract DE-SC0011385. Yi Jiang is supported by DOE grant DE-FG02-11ER16210. Elliot Padgett is supported by a NSF Graduate Research Fellowship (DGE-1144153).